\pdfoutput=1
\documentclass[11pt]{article}

\usepackage[final]{acl}

\usepackage{times}
\usepackage{latexsym}
\usepackage{array} % Required for column type definition
\usepackage{amssymb}   % For checkmark symbol
\usepackage{pifont}% http://ctan.org/pkg/pifont
\newcommand{\cmark}{\ding{51}}%
\newcommand{\xmark}{\ding{55}}%
\usepackage[T1]{fontenc}
\usepackage{stfloats}
\usepackage[utf8]{inputenc}
\usepackage{dev}

\usepackage{microtype}

\usepackage{inconsolata}

\usepackage{graphicx}
\usepackage{rotating}
\usepackage{makecell}

\title{Bhaasha, Bhāṣā, Zaban: A Survey for Low-Resourced Languages in South Asia – Current Stage and Challenges}

\author{Sampoorna Poria\textsuperscript{1}, Xiaolei Huang\textsuperscript{2} \\
  \textsuperscript{1} Dept of Computer Science \& Engineering, West Bengal University of Technology,\\
  \textsuperscript{2} Department of Computer Science, University of Memphis\\
  \href{mailto:sampoornap0172@gmail.com}{sampoornaporia@gmail.com}, \href{mailto:xiaolei.huang@memphis.edu}{xiaolei.huang@memphis.edu}}

\begin{document}

\maketitle
\begin{abstract}
Rapid developments of large language models have revolutionized many NLP tasks for English data. Unfortunately, the models and their evaluations for low-resource languages are being overlooked, especially for languages in South Asia.
Although there are more than 650 languages in South Asia, many of them either have very limited computational resources or are missing from existing language models.
Thus, a concrete question to be answered is: \textit{Can we assess the current stage and challenges to inform our NLP community and facilitate model developments for South Asian languages?}
In this survey\footnote{Bhaasha (Hindi), Bhāṣā (Bengali), and Zabān (Urdu/Persian) all mean ``language'' and are commonly used across South Asian language families, underscoring the paper's inclusive focus.}, we have comprehensively examined current efforts and challenges of NLP models for South Asian languages by retrieving studies since 2020, with a focus on transformer-based models, such as BERT, T5, \& GPT. 
We present advances and gaps across 3 essential aspects: data, models, \& tasks, such as available data sources, fine-tuning strategies, \& domain applications. 
Our findings highlight substantial issues, including missing data in critical domains (e.g., health), code-mixing, and lack of standardized evaluation benchmarks.
Our survey aims to raise awareness within the NLP community for more targeted data curation, unify benchmarks tailored to cultural and linguistic nuances of South Asia, and encourage an equitable representation of South Asian languages. The complete list of resources is available at: \href{https://github.com/trust-nlp/LM4SouthAsia-Survey}{https://github.com/trust-nlp/LM4SouthAsia-Survey}.\footnote{This work was done when the first author was a remote intern at the University of Memphis.}
\end{abstract}

% -------------------------------------------------------

\section{Introduction}

% While a quarter of the world's population resides in this region \citep{veron2008demography}, its languages remain severely underrepresented in NLP models \citep{dunn-etal-2024-pre}.

South Asia is one of the most linguistically diverse regions, encompassing Indo-Aryan, Dravidian, Iranian, and Tibeto-Burman languages, along with numerous isolates \citep{arora-etal-2022-computational, borin-etal-2014-linguistic}.
However, the regional languages are often missing from training corpora or present in imbalanced quantities \citep{Khan_2024}, and many of them are not supported by current large language models (LLMs) \citep{lai-etal-2024-llms}.
There are multiple factors behind this disparity, and it's crucial to identify and address them to ensure better representation of South Asian languages.
The definition of ``low-resource'' varies based on data availability and digital presence \citep{nigatu-etal-2024-zenos,
mehta-etal-2020-learnings}. 
We consider a language ``low-resource'' if it lacks computational data and standardized evaluation benchmarks for most NLP tasks. 
Crucially, this framing moves beyond definitions based solely on speaker population, since even widely spoken languages like Hindi and Bengali remain under-resourced in terms of benchmark coverage and model support.
While low-resource languages have been studied for various regions \citep{aji-etal-2023-current, aji-etal-2022-one, adebara-abdul-mageed-2022-towards}, there is no comprehensive study on the current status of South Asian NLP, which will be fulfilled by this survey, as outlined in Table~\ref{tab:comparison}.

\paragraph{Study retrieval methods.} We retrieved relevant studies from 2020 onward via ACL Anthology, Semantic Scholar, and Google Scholar by broad and specific keyword combinations.
We extended the publication list by screening their citation networks in Google Scholar, such as journals or workshop venues.
To assess on the latest trends, we excluded papers before 2020 and focused on neural and Transformer-based models. The detailed methodology is presented in Appendix~\ref{sec:appendix-retrieval}.

\begin{table}
    \centering
    \resizebox{.5\textwidth}{!}{
    \begin{tabular}{c|ccccc}
    \makecell{Study} & \makecell{Inclusive\\Language\\Coverage} & \makecell{Data\\Insights} & \makecell{Multiple\\NLP Tasks} & \makecell{Interdisciplinary\\Integration} & \makecell{Recent\\LLMs}\\
    \hline\hline
    \citeauthor{hedderich-etal-2021-survey} & \textbf{\cmark$^1$} & \textbf{\cmark} & \textbf{\cmark} & \textbf{\xmark} & 
    \textbf{\xmark} \\
    \citeauthor{arora-etal-2022-computational} & \textbf{\cmark} & \textbf{\cmark} & \textbf{\cmark} & \textbf{\cmark} & \textbf{\xmark$^2$}\\
    \citeauthor{10.1145/3695766} & \textbf{\xmark} & \textbf{\cmark} & \textbf{\cmark} & \textbf{\xmark} & \textbf{\xmark}\\
    \citeauthor{10.1145/3567592} & \textbf{\cmark$^1$} & \textbf{\cmark} & \textbf{\xmark} & \textbf{\xmark} & \textbf{\xmark$^3$}\\
    Our Work & \textbf{\cmark} & \textbf{\cmark} & \textbf{\cmark} & \textbf{\cmark} & \textbf{\cmark}
    \end{tabular}
    }
  \caption{\label{tab:comparison}
    Comparing related surveys of low-resourced languages to ours by multiple key criteria. We denote superscript $^1$ as not specific to South-Asian languages; $^2$ as limited discussion of LLMs; and $^3$ as related to multilingual models but not for LLMs or low-resourced languages. ``Interdisciplinary Integration'' refers to studies connecting NLP with health, education, etc. 
  }
\end{table}

\paragraph{Objectives and Contributions.}
We assess the current state of NLP research for South Asian languages and summarize their key issues, evaluation limits, and research gaps unique to these languages. 
Unlike prior related surveys in Table~\ref{tab:comparison}, our work makes three unique contributions: 1) we present comprehensive language families in South Asia and broadens coverage beyond Indo-Aryan and Dravidian languages by covering other widely spoken language families in the region; 
2) we examine data sources and provide data insights to accelerate low-resourced language research in South Asia;
and 3) we analyze studies across various domains (e.g., healthcare and education) and summarize recent LLMs and their tuning strategies (e.g., LoRA~\citep{hu2021loralowrankadaptationlarge}).
We hope this survey will inspire future directions to strengthen NLP community efforts for underrepresented languages in South Asia.

\begin{table*}[!t]
    \centering
    % Adjust font size if needed. For example: \tiny, \scriptsize, \footnotesize, etc.
    \scriptsize 
    % Resizes the entire table to fit the page width
    \resizebox{\textwidth}{!}{%
    \begin{tabular}{llllllll}
    \textbf{Data}  & \textbf{Language(s)}  & \textbf{Size}  & \textbf{NLP Task}  & \textbf{Year}  & \textbf{Source}  & \textbf{Domain}  & \textbf{Acc} \\
    \hline
    \multicolumn{8}{c}{\textbf{Datasets}} \\
    \hline
    \hypertarget{INDIC-MARCO}{INDIC-MARCO} & Multiple (11) & 8.8M & Neural IR & 2024 & \citeauthor{Haq2023} & General & Yes\\
    \hypertarget{BPCC}{BPCC} & Multiple( 22) & 230M & Machine Translation & 2023 & \citeauthor{Gala2023} & General & Yes \\
    TransMuCoRes & Multiple (31) & 1.8M & Coreference Resolution & 2024 & \citeauthor{Mishra2024} & General & Yes\\
    Samanantar & Multiple (11) & 12.4M & Machine Translation & 2022 & \citeauthor{ramesh2022samanantar} & General & Yes \\
    IndicCorp & Multiple (11) & 453M & LM Pretraining & 2020 & \citeauthor{kakwani-etal-2020-indicnlpsuite} & News & Yes\\
    Sangraha & Multiple (22) & 74.8M & LM Pretraining & 2024 & \citeauthor{Khan_2024} & General & Yes \\
    HinDialect & Multiple (26) & - & Model Pretraining & 2022 & \citeauthor{bafna-etal-2022-combining} & General & Yes \\
    L3Cube-IndicNews & Multiple (11) & 360K & Headline/Document Classification & 2023 & \citeauthor{mirashi2024l3cubeindicnewsnewsbasedshorttext} & News & Yes\\
    Aksharantar & Multiple(21) & 26M & Transliteration & 2023 & \citeauthor{madhani-etal-2023-aksharantar} & General & Yes\\
    \hypertarget{PMIndia}{PMIndiaSum} & Multiple (14) & 697K & Multilingual Summarization & 2023 & \citeauthor{Urlana2023} & Government & Yes\\
    CVIT-PIB v1.3 & Multiple(11) & 2.78M & Multilingual NMT & 2021 & \citeauthor{Philip2021} & Government & Yes\\
    IndicSynth & Multiple (12) & 4000 & Audio Deepfake Detection & 2025 & \citeauthor{sharma-etal-2025-indicsynth} & General & Yes\\
    CaLMQA & Multiple (23) & 1.5K & LFQA & 2024 & \citeauthor{arora2024calmqa} & Culture\&Society & Yes\\
    MultiCoNER & Multiple (11) & 26M & NER & 2022 & \citeauthor{Malmasi2022} & Wiki\&Search & Yes\\
    Homophobia Data & Telugu, Kannada, Gujarati & 38,904 & Homophobia Detection & 2024 & \citeauthor{Kumaresan2024} & Social Media & No\\
    Fake News Detection & Malayalam & 1,682 & Fake News Detection/ Classification & 2024 & \citeauthor{K2024} & News Media & No\\
    POS Tagging Dataset & Angika, Magahi, Bhojpuri & 2124 & POS tagging & 2024 & \citeauthor{Kumar2024} & News,Conversations & Yes\\
    Assamese BackTranslit & Assamese & 60K & Back transliteration & 2024 & \citeauthor{Baruah2024} & Social Media & Yes \\
    IruMozhi & Tamil & 1,497 & Diglossia Classification & 2024 & \citeauthor{Prasanna2024} & Wikipedia & Yes \\
    Paraphrase Corpus & Pashto & 6,727 & Paraphrase detection & 2024 & \citeauthor{Ali2024} & News Media & Yes\\
    Hate Speech Data & Bengali, Hindi, Urdu & - & Sentiment Analysis, Hate Detection & 2024 & \citeauthor{Hasan2024} & Social Media & No\\
    AS-CS Dataset & Hindi, Bengali & 5,062 & Counter Speech Generation & 2024 & \citeauthor{Das2024} & Social Media & Yes\\
    CoPara & 4 Dravidian Languages & 2856 & Paragraph-level alignment & 2023 & \citeauthor{E2023} & News Media & Yes\\
    
    NP Chunking Data & Persian & 3,091 & Noun Phrase Chunking & 2022 & \citeauthor{Kavehzadeh2022} & News Media & No\\
    Punctuation Dataset & Bengali & 1.3M & Punctuation Restoration & 2020 & \citeauthor{Alam2020} & News\&Stories & Yes\\
    \hypertarget{L3Cube-MahaCorpus}{L3Cube-MahaCorpus} & Marathi & 289M & Classification \& NER & 2022 & \citeauthor{Joshi2022}. & News/Non-news & Yes\\
    HATS & Hindi & 405 & LLM Reasoning Evaluation & 2025 & \citeauthor{gupta-etal-2025-hats} & Education & Yes \\
    WoNBias & Bengali & 31,484 & Bias Classification & 2025 & \citeauthor{aupi-etal-2025-wonbias} & Culture\&Society & Yes\\
    UFN2023 & Urdu & 4,097 & Human/Machine Fake News Detection & 2025 & \citeauthor{ali-etal-2025-detection} & News & Yes\\
    Flickr30K (EN-(hi-IN)) & Hindi & 156,915 & Multimodal Machine Translation & 2018 & \citeauthor{Chowdhury2018} & Image Captions & Req\\
    \hypertarget{SENTIMOJI}{SENTIMOJI} & Hindi & 20k & Emoji Prediction & 2024 & \citeauthor{Singh2024} & Social Media & Yes \\
    Suman & Kadodi,Marathi & 942 & Machine Translation & 2024 & \citeauthor{Dabre2024} & Conversation & Yes\\
    WMT24 En-Hi Data & Hindi & 1500 & Machine Translation & 2024 & \citeauthor{Bhattacharjee2024} & Mutlidomain & Yes\\
    AGhi & Hindi & 36,670 & AI-generated text detection & 2024 & \citeauthor{Kavathekar2024} & News & Yes\\
    Mizo News Summarization Dataset & Mizo & 500 & News Summarization & 2024 & \citeauthor{Bala2024} & News & Yes\\
    ADIhi & Hindi & 36,670 & Ranking LLMs on AI Detectability & 2024 & \citeauthor{Kavathekar2024} & News & Yes\\
    En-Tcy test dataset & Tulu & 1300 & Machine Translation & 2024 & \citeauthor{Narayanan2024} & Wiki,FLORES & Yes\\
    \hypertarget{MMCQS dataset}{MMCQS dataset} & Hindi & 3,015 & Multimodal Ques. Summarization & 2024 & \citeauthor{Ghosh2024} & Healthcare & Yes\\
    
    BNSENTMIX & Bengali & 20K & Sentiment Analysis & 2025 & \citeauthor{alam-etal-2025-bnsentmix} & Social Media & Yes\\
    VACASPATI & Bengali & 11M & Multiple Downstream Tasks & 2023 & \citeauthor{bhattacharyya-etal-2023-vacaspati} & Literature & Yes \\
    Multi³Hate & Hindi & 300 & Multimodal Hate Detection & 2025 & \citeauthor{bui-etal-2025-multi3hate} & Social Media & Yes\\

    MDC³ & Bengali & 5,007 & Commercial Content Classification & 2025 & \citeauthor{shanto-etal-2025-mdc3} & Social media & Yes\\

    Hindi-BEIR & Hindi & 5.89M & 7 Retreival Tasks & 2025 & \citeauthor{acharya-etal-2025-benchmarking} & General & Yes\\
    % \hline
    \hline
    \multicolumn{8}{c}{\textbf{Benchmarks}} \\
    \hline

    IN22 Benchmark & Multiple (22) & 2527 & Machine Translation & 2023 & \citeauthor{Gala2023} & General & Yes\\
    BELEBELE & Multiple (122 variants) & 900 & Multilingual Reading Comp. & 2024 & \citeauthor{Bandarkar2024} & Web Articles & Yes\\
    
    Multilingual DisCo & Multiple(6) & 84 & Geneder Bias Evaluation & 2023 & \citeauthor{Vashishtha2023} & General & Yes\\
    
    IndicNLG Benchmark & Multiple (11) & 8.5M & Various Generative Tasks & 2022 & \citeauthor{Kumar2022} & News, Wiki & Yes\\
    
    IndicGlue & Multiple (11) & 2M & Various NLU Tasks & 2020 & \citeauthor{kakwani-etal-2020-indicnlpsuite} & News, Wiki & Yes\\
    
    Indic-QA & Multiple (11) & - & LLM Q\&A Capabilities & 2025 & \citeauthor{singh-etal-2025-indic} & General & Yes\\

    MILU & Multiple (11) & 79,617 & Knowledge/Reasoning Evaluation & 2025 & \citeauthor{verma-etal-2025-milu} & Multiple & Yes\\

    En-Hi Chat Translation & Hindi & 16,249 & Chat Translation & 2022 & \citeauthor{Gain2022} & Customer Service & Yes\\
    
    CounterTuringTest(CT2) & Hindi & 26 & Benchmarking AGTD techniques & 2024 & \citeauthor{Kavathekar2024} & News & Yes\\
    
    MMFCM & Hindi & - & Multimodal Ques. Summarization & 2024 & \citeauthor{Ghosh2024} & Healthcare & Yes\\

    BenNumEval & Bengali & 3.2k & LLM Numerical Reasoning Capabilities & 2025 & \citeauthor{ahmed-etal-2025-bennumeval} & Yes\\
    
    \end{tabular}%
    }
    \caption{\label{tab:data}
    Available Datasets and Benchmarks for Low-Resource South Asian Languages Across Tasks and Domains,  organized by resource type (task-specific and general-purpose datasets, followed by benchmarks). We denote `Req' as Available on Request; `Acc' as Public Accessibility.}
\end{table*}

\begin{figure*}[ht]
    \centering
    \includegraphics[width=0.75\textwidth]{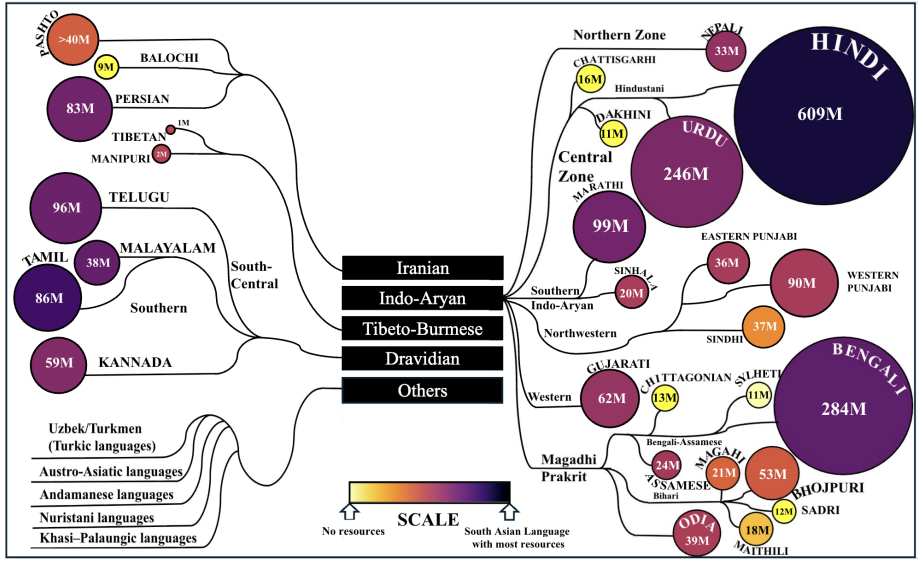}  % Ensures it fits within two columns
    \caption{Language families regarding Speaker population and Resource availability. Bubble Size indicates speaker population per language and color intensity indicates the amount retrieved NLP resources. Darker color means more resources, and vice versa. "Resource size" refers to the number of papers in the ACL Anthology (until 2024) that mention the language in the title and/or abstract. Languages primarily spoken outside South Asia (e.g., Uzbek) are excluded from resource size visualization to maintain regional focus.}
    \label{fig:language_distribution}
\end{figure*}

\section{Data and Resources}

A large text corpus is essential to enable language models to understand complex and heterogeneous semantics and structures of South Asian languages.
Indeed, over 650 languages are spoken in the region, yet computational resources remain scarce and highly skewed toward a few languages \citep{zhao-etal-2025-enhancing, Hasan2024, Narayanan2024, Ali2024, Baruah2024}.
For example, most language resources consist of small text samples, with a major focus on languages like Hindi and Urdu \citep{kakwani-etal-2020-indicnlpsuite,Philip2021, Gala2023}.
However, existing studies may merely address the questions that will be answered in our study:
1) \textit{What are the available corpora for the low-resourced languages in South Asia?}
2) \textit{What NLP tasks are in the corpora?}
and 3) \textit{What domains are the corpora?}
To answer those questions, we summarize data distributions by language families in Figure~\ref{fig:language_distribution} and statistics in Table~\ref{tab:data}.

\subsection{Language resources} \label{sec:language_resources}
Figure \ref{fig:language_distribution} presents the uneven distribution of South Asian languages in our collected resources. 
The color gradient and circle sizes show that there are a few dominant languages with comparatively more resources, such as Hindi, Bengali, and Telugu, while the others are severely underrepresented. This highlights resource challenges and opportunities.
We categorize retrieved studies by language family: Indo-Aryan, Dravidian, Tibeto-Burman, and Iranian languages.

\paragraph{Indo-Aryan Languages} own the largest language population in South Asia and are relatively more represented in our collected studies. 
For example, Hindi, Bengali, Marathi, and Urdu are among the largest bubbles in Figure~\ref{fig:language_distribution}, and Hindi corpora are available for all major NLP tasks in Table~\ref{tab:data}, aligning with existing language speaker populations~\citep{Gala2023}.
Large-scale data are not evenly-distributed across NLP tasks.
For instance, IndicMARCO, IndicCorp, IndicGlue, MultiCONER, and BELEBELE offer large-scale datasets for IR, model pretraining, NER, and reading comprehension, particularly in high-resource Indic languages \citep{Haq2023, Malmasi2022, Bandarkar2024,kakwani-etal-2020-indicnlpsuite}.
However, Bhojpuri, Sindhi, and Assamese are only in a few domain-specific datasets \citep{Baruah2024, Malmasi2022, Kumar2024}: their dataset size is comparatively smaller (less than 5,000 samples) \citep{Gala2023}.
% HinDialect provides monolingual corpora for 26 dialects of the Hindi belt, many previously unsupported (Bafna et al., 2022).
% Multiple code-mixed data of sentiment and emoji prediction tasks are available for Hindi, Bengali, and Marathi (e.g., MMCQS, SENTIMOJI, and BSENTMIX) \citep{Singh2024, Ghosh2024, alam-etal-2025-bnsentmix}.

\paragraph{Dravidian Languages} include Tamil, Malayalam, Telugu, and Kannada in a number of integrated multilingual corpora \citep{Gala2023, Haq2023, Urlana2023, Philip2021, mirashi2024l3cubeindicnewsnewsbasedshorttext} for NLP tasks, such as diglossia classification, machine translation, and hate speech detection \citep{Prasanna2024, Kumaresan2024, K2024}.
However, many Dravidian languages, including Kodava, Toda, and Irula, are absent from major data resources and benchmarks. 
A rare exception is Tulu, which is included in a recently developed parallel corpus for machine translation \citep{Narayanan2024}.
The language resources are relatively smaller in size compared to Indo-Aryan Languages (e.g., Hindi) and cover much fewer application domains, such as healthcare.

\paragraph{Tibeto-Burman and Iranian Languages} are critically underrepresented. South Asia is home to 245 Tibeto-Burman and 84 Iranian languages \citep{glottolog, ethnologue}, yet only a handful resource appear in available datasets. Manipuri, Mizo, and Bodo are \textbf{Tibeto-Burman languages} in our retrieved studies, such as
summarization data \citep{Urlana2023, Bala2024, madhani-etal-2023-aksharantar}. However, the other languages including Dzonkgkhe (the national language of Bhutan) are not covered. \textbf{Iranian Languages} including Pashto, Persian, \& Balochi are available in our data collections, such as a paraphrase detection corpus in Pashto \citep{Ali2024}, a noun phrase chunking corpus in Persian \citep{Kavehzadeh2022}, and a question answering corpus in Balochi \citep{arora2024calmqa}.
While IndicNLG is one of the largest benchmarks, many Tibeto-Burman and Iranian languages (e.g., Dari \& Wakhi) are largely missing~\citep{Kumar2022}. 

\subsection{NLP Tasks}

The availability of NLP tasks varies by language in Table~\ref{tab:data}.
For example, Indo-Aryan languages cover all major NLP tasks, such as machine translation, information extraction, and sentiment analysis; in contrast, the other language families only cover very few NLP tasks.
This section summarizes major NLP tasks from the data perspective in two major categories, 1) \textit{generative} and 2) \textit{discriminative} tasks. Methodologies are referred to in Section~\ref{sec:model}.

\paragraph{Generative NLP tasks} cover three major tasks, machine translation, text generation, and summarization. \underline{Machine translation} is the most represented task in Table~\ref{tab:data}, including BPCC \citep{Gala2023} and domain-specific parallel corpora CVIT-PIB v1.3 and Suman \citep{Philip2021, Dabre2024}. 
However, Kashmiri, Sindhi, and Tulu lack sufficient bilingual corpora––relying on back-translation \citep{Baruah2024} and cross-lingual transfer \citep{Narayanan2024}.
The scarcity of consistent annotations and high-quality datasets can be a critical issue.
\underline{Text Summarization} is mainly in general domains (e.g., news) for Indo-Aryan languages, such as PMIndiaSum \citep{Urlana2023}, \& misses coverages of Dravidian and Tibeto-Burman languages.
MedSumm data aids in multimodal summarization for Hindi-English code-mixed clinical queries, specifically for the healthcare \citep{Ghosh2024}, while domain-specific summarizations are not available in other languages.
\underline{Text Generation} resources include the IndicNLG benchmark \citep{kumar-etal-2022-indicnlg}, which covers biography generation, news headline generation, sentence summarization, paraphrasing, \& question generation across 11 Indic languages. Long-form question answering remains underdeveloped \citep{arora2024calmqa}, \& chat translation resources are also scarce \citep{Gain2022}

\paragraph{Discriminative NLP tasks} mainly focus on sequential classifications, such as Named entity recognition (NER).
Classification tasks account for the majority of discriminative NLP tasks in our study, such as sentiment analysis \& hate speech detection. 
For example, SENTIMOJI (sentiment prediction data for Hindi-English code-mixed texts) \citep{Singh2024}, and hate detection resources are available for Hindi, Tamil, Bengali, \citep{Hasan2024}, Kannada, and Telugu \citep{K2024}. 
However, sentiment analysis \& hate speech detection data remain nearly absent for Tibeto-Burman \& Iranian languages.
The table also shows that semantic or syntactic tasks are most likely available for Hindi, such as syntactic parsing \& coreference resolution~\citep{Kumar2024, Mishra2024}.
% While POS tagging and syntactic parsing has been  well explored in English, the tasks are only available with small sizes for Universal Dependencies (UD) parsing \citep{Baruah2024} and some low-resource Indo-Aryan languages such as Angika, Magahi, and Bhojpuri~\citep{Kumar2024}. 
% Coreference resolution datasets are almost nonexistent, and limited resources are available for Hindi, where most datasets focus on sentence-level tasks rather than document-level coherence modeling \citep{Mishra2024}. 
Similarly, recently new data releases are primarily for Hindi, such as AI-generated text detectability ~\citep{Kavathekar2024}.

% \textit{While Table 2 shows extensive resources for tasks like information retrieval and named entity recognition [], many other important tasks like coreference resolution and discourse analysis have less resources even for languages receiving more attention. Most multilingual datasets are concentrated in generic domains like news or web-articles with limited coverage for critical areas like healthcare and education. Additionally, only a few benchmarks () are large-scale, while the majority comprise fewer than 5,000 examples. This imbalance results in challenges for model training and evaluation for less-represented tasks and domains. }

% Three subsections
% Current Multlingual Models
% Adaptation or Fine-tune approaches, Lora or other techniques.
% Model evaluations
% Trends and Challenges (e.g., tokenizers)

\begin{table*}[htp]
  \centering
  \resizebox{\textwidth}{!}{
  \begin{tabular}{llllccc}
    
    \textbf{Model}  & \textbf{Architecture}  & \textbf{Language}  & \textbf{Training Strategy}  & \textbf{Parameter Size}  & \textbf{Year}  &
    \textbf{Source}\\
    \hline
    AxomiyaBERTa & BERT & Assamese & Continuous Pre-train + Supervised Fine-tuning & 66M & 2023 & \citeauthor{Nath2023}\\
    
    IndecBERT & BERT & Multiple (11) & Continuous Pre-train on IndicCorp + Supervised Fine-tuning & 12M & 2020 & \citeauthor{kakwani-etal-2020-indicnlpsuite}\\

    IndicBART & BART & Multiple (11) & Continuous Pre-train on IndicCorp + Supervised Fine-tuning & 244M & 2022 & \citeauthor{Dabre_2022}\\
    
    BUQRNN & LSTM+BERT & Bengali & Supervised Training & NA & 2024 & \citeauthor{Yu2024}\\
    
    PN-BUQRNN & LSTM+BERT & Bengali & Supervised Training & NA & 2024 & \citeauthor{Yu2024}\\

    Matina & Transformer & Persian & Domain-specific Fine-tuning & 8B & 2025 & \citeauthor{hosseinbeigi-etal-2025-matina-culturally}\\

     IndicTrans & Transformer & Multiple (11) & Continuous Pre-train on Samanatar + Supervised Fine-tuning & 1.1B & 2022 & \citeauthor{ramesh2022samanantar}\\
    
    IndicTrans2 & Transformer & Multiple (22) & Pre-train + Supervised Fine-tuning & 1.1B & 2023 & \citeauthor{Gala2023}\\
    
    DC-LM & BERT & Kannada & Supervised Fine-tuning & 110M & 2022 & \citeauthor{Hande2022}\\
    
    Lambani NMT & Transformer & Lambani & Pre-train + Supervised Fine-tuning & 380M & 2022 & \citeauthor{Chowdhury2022}\\
   
    Indic-ColBERT & BERT & Multiple (11) & Supervised Fine-tuning
 & 42M & 2023 & \citeauthor{Haq2023}\\

    MedSumm & Multiple LLMs & Hindi (Code-mixed) & Supervised Fine-tuning & 7B-13B & 2024 & \citeauthor{Ghosh2024}\\
    
    Tri-Distil-BERT & BERT & Bengali, Hindi & Continuous Pre-train & 8.3B & 2024 & \citeauthor{raihan2023mixed}\\
    
    Mixed-Distil-BERT & BERT & Bengali, Hindi & Continuous Pre-train + Supervised Finetuning & 8.3B & 2024 & \citeauthor{raihan2023mixed}\\
   
    CPT-R & Llama  & Multiple (5) & Continuous Pre-train & 7B & 2024 & \citeauthor{j-etal-2024-romansetu}\\
  
    IFT-R & Llama & Multiple (5) & Instruction Fine-tuning  & 7B & 2024 & \citeauthor{j-etal-2024-romansetu}\\
 
    BASE & GRU & Hindi & Supervised Training & NA & 2023 & \citeauthor{Daisy2023}\\

    MED & Bi-GRU & Hindi & Supervised Training & NA & 2023 & \citeauthor{Daisy2023}\\
  
    RETRAIN& Bi-GRU & Hindi & English Gigaword Pre-train + Supervised Fine-tuning & NA & 2023 & \citeauthor{Daisy2023}\\
    
    % DT-GNN-enhanced Transformer	& BERT & Hindi, Bengali & Leipzig Corpora Collection & 110M & 2022 & \citeauthor{Venkatesh2022}\\
    
    Nepali DistilBERT & BERT & Nepali & Nepali corpora Pre-train by Progressive Mask & 66M & 2022 & \citeauthor{Maskey2022}\\
 
    Nepali DeBERTa & BERT & Nepali & Nepali Corpora Pre-train by Mask-LM & 110M & 2022 & \citeauthor{Maskey2022}\\
   
    TPPoet & Transformer & Persian & Persian poetry Pretrain + Supervised Fine-tuning & 33M & 2023 & \citeauthor{Panahandeh2023}\\

    MahaBERT & BERT & Marathi & L3Cube-MahaCorpus Pre-train  & 110M & 2020 & \citeauthor{Joshi2022}\\

    Emoji Predictor & Transformer & Hindi (Code-mixed) & Supervised Fine-tuning & NA & 2024 & \citeauthor{Singh2024} \\
  
    RelateLM & BERT & Multiple (5) & Wiki/CFILT Pre-train + Supervised Fine-tuning & 110M & 2021 & \citeauthor{Khemchandani2021}\\
   
    Multi-FAct & Mistral-7B & Bengali & Supervised Fine-tuning & 7B & 2024 & \citeauthor{Shafayat2024} \\

    AI-Tutor & Transformer & Pali, Ardhamagadhi & Pre-train + Supervised Training & 1.1B & 2024 & \citeauthor{Dalal2024}\\

    LlamaLens & Transformer & Hindi & Instruction tuning + Domain Fine-tuning; Multilingual Shuffling & 8B & 2025 & \citeauthor{kmainasi-etal-2025-llamalens}\\

    NLLB-E5 & Multilingual Encoder & Hindi & Knowledge Distillation + Zero-shot transfer & 1.3B & 2025 & \citeauthor{acharya-etal-2025-benchmarking}\\

  \end{tabular}
  }
  \caption{\label{tab:model}
    Model summary by language, architecture, training strategies, and others. 
  }
\end{table*}

\section{Model Advances}
\label{sec:model}

We examine recent model advances of South Asian languages in Table~\ref{tab:model} — covering three major topics, multilingual language models, training and fine-tuning methods, and model evaluations.

\subsection{Multilingual Language Models}

\paragraph{Code-Mixed Tokenization} is the fundamental step to encode input text containing characters from multiple languages and usually starts by fine-tuning existing language model tokenizers.
For example, \citet{kumar-etal-2023-indisocialft} train FastText~\citep{bojanowski-etal-2017-enriching} on code-mixed, transliterated, and native-script social media text for multiple Indic languages, other studies fine-tune BERT~\citep{devlin2019bert} or multilingual BERT tokenizers to predict positive hope speech in Kannada-English \citep{Hande2022}, Hindi-English sentiments \citep{Singh2024}, and review ratings \citep{Yu2024}.
% , followed by quantum recurrent neural networks (QRNNs).
The Overlap BPE method \citep{Patil2022} improves tokenization consistency on subword-level processing for orthographically similar languages. 

\paragraph{Transformer-based models}~\citep{vaswani2017attention} have dominated recent developments for monolingual and multilingual settings.
BERT is a common architecture on multi-domain and monolingual tasks, such as AxomiyaBERTa \citep{Nath2023}, Nepali DistilBERT and DeBERTa \citep{Maskey2022}, and MahaBERT \citep{Joshi2022}. 
For multilingual models, IndicBERT \citep{kakwani-etal-2020-indicnlpsuite} covers classification and retrieval;
IndicTrans2 \citep{Gala2023} covers translation across 22 languages;
Indic-ColBERT \citep{Haq2023} employs retrieval-augmented supervision for search to improve document retrieval across 11 languages;
and IndicBART \citep{Dabre_2022} supports NMT \& summarization across 2 language families. Together, these represent some of the most comprehensive models for South Asian languages.
\citet{Chowdhury2022} trains Transformer models from scratch for machine translation to Lambani, using data from closely related source languages.
Classification tasks mainly use supervised fine-tuning on pre-trained BERT~\citep{devlin2019bert} and its variants.

% \paragraph{Large language models (LLMs)} 
Generative LLMs are being rapidly adopted for South Asian languages in the recent 3 years.
MedSumm \citep{Ghosh2024} fine-tuned 5 public LLMs (Llama 2~\citep{touvron2023llama2openfoundation}, FLAN-T5~\citep{chung2022scalinginstructionfinetunedlanguagemodels}, Mistral~\citep{jiang2023mistral7b}, Vicuna~\citep{zheng2023judging}, and Zephyr~\citep{tunstall2024zephyr}) on medical question summarization with visual cues for code-mixed Hindi-English patient queries.
Multi-FAct \citep{Shafayat2024} uses Mistral-7B \citep{jiang2023mistral7b} to extract facts from LLM-generated texts.
CPT-R and IFT-R \citep{j-etal-2024-romansetu} fine-tuned LLaMA2-7B models on romanized Indic corpora to enable transliteration-aware and mixed-script text processing. 
Additionally, AI-Tutor \citep{Dalal2024} applied IndicTrans2 \citep{Gala2023} to Pali and Ardhamagadhi.
% Due to LLM sizes, fine-tuning partial parameters are more practical.
These findings suggest that multilingual models alone cannot resolve low-resource challenges in South Asia; corpus coverage and script fidelity continue to constrain their applicability, particularly for languages with limited web presence and domain coverage.

\subsection{Training and Fine-tuning Methods}

\paragraph{Code-mixed and script-specific adaptations} enable model understanding of text inputs with mixed languages.
For example, LLMs struggled with Bengali script generation due to inefficient tokenization~\citep{Mahfuz2024}.
Studies introduced related corpora to assess code-mixed capabilities, such as IndicParaphrase \citep{kumar-etal-2022-indicnlg}, the largest Indic language paraphrasing dataset across 11 languages.
% , highlight that script-specific adaptations enhance paraphrase generation and improve multilingual understanding in code-mixed settings. 
% \citet{Huzaifah2024} found the large model and diverse data may help code-switching translation that GPT-4 and GPT-3.5 outperform traditional NMT models.
Transliterating Indic languages into a common script could effectively improve cross-lingual transfer, such as NER and sentiment analysis~\citep{Moosa2022}.
\citet{Kirov2024} aligned transliteration patterns with phonetic structures, which further improves multilingual representation.
Overlap BPE \citep{Patil2022} finds shared subword representations, which enhances consistency for orthographically similar languages.
Continual pre-training strategies \citep{guo-etal-2025-efficient-domain, zheng-etal-2024-breaking} improve adaptation without degrading prior performance, for example in machine translation \citep{Koehn2024}, by preventing catastrophic forgetting by iteratively fine-tuning with new language pairs. 
\citet{agarwal-etal-2025-script} introduces script-agnostic representations for Dravidian languages and show that mixing multiple writing systems during training improves robustness.
While the current studies have achieved substantial progresses, script-aware tokenization remains a foundational bottleneck to enable encoding multilingual inputs of South Asian languages.

\paragraph{Supervised multilingual transfer learning}
Given the linguistic similarities in characters and morphology, cross-lingual transfer learning has become a key adaptation strategy. \citet{Narayanan2024}, IndicBART \citep{Dabre_2022}, and IndicTrans2 \citep{Gala2023} show that pre-training on large multilingual corpora of related languages (that can be mapped to a single script) significantly improves translation.
Llama 2-based models \citep{j-etal-2024-romansetu} were fine-tuned on task-specific corpora; however, effectiveness varies based on linguistic proximity, with underrepresented languages facing performance declines \citep{Hasan2024}.
Studies found that jointly trained NER models on multilingual corpora outperformed monolingual ones as for shared script and grammar, such as Hindi-Marathi \citep{Sabane2023} and Bengali-Tamil-Malayalam \citep{Murthy2018}. 
% Back-translation and synthetic data augmentation improved Hindi, Sinhala, and Tamil NMT \citep{Chowdhury2018}. 

Several studies explored finetuning approaches.
Adaptive multilingual finetuning \citep{Das2023} leverages subword embedding alignment to enhance transferability across related languages.
\citet{Zhou2023} integrates sociolinguistic factors into offensive language detection.
\citet{Poudel2024} fine-tunes with domain-specific knowledge to enhance legal translation.
Cross-lingual in-context learning (ICL) \citep{Cahyawijaya2024} improve generalization by query alignment.

\paragraph{Distillation and parameter-efficient finetuning (PEFT) methods} Adapting large models to South Asian languages often face computing and data constraints.
As a result, recent work has explored PEFT strategies like LoRA, QLoRA, and multi-step PEFT \citep{hu2021loralowrankadaptationlarge, Petrov2023}. These approaches fine-tune models like Gemma \citep{khade-etal-2025-challenges} with fewer parameters and lower memory cost.
While LoRA improves efficiency, its effectiveness can vary across tasks: it captures dialectal variations when combined with phonological cues \citep{alam-anastasopoulos-2025-large} but may struggle with syntactically rich tasks.
Adapter-based methods \citep{Nag2024} offer modular, language-specific adaptation and can avoid catastrophic forgetting when tuned with domain/task-specific knowledge. 

Distillation-based approaches \citep{Ghosh2024} compress large models but typically require access to high-quality teacher models and synthetic data, which remains a bottleneck in many South Asian contexts.
Feature-based finetuning \citep{Bhatt2022} focuses on internal representation refinement to enable knowledge transfer across resource boundaries.
Other strategies like rank-adaptive LoRA \citep{Yadav2024} balance parameter savings with performance.
Complementary strategies such as QLoRA \citep{dettmers2023qlora} reduce memory overhead, while data-centric approaches like IndiText Boost \citep{Litake2024} combine augmentation techniques to enhance classification for morphologically rich languages (e.g., Sindhi, Marathi).
Few-shot learning offers flexibility but still struggles with syntactic generalization \citep{Nag2024, Pal2024}.
While parameter-efficient \& data-light methods have achieved progress, their benefits are uneven across linguistic variations, rarely extending to the least-resourced.

\subsection{Model Evaluations}
Model evaluation varies by task, such as BLEU for generation and human evaluation \citep{Gala2023, Narayanan2024, duwal-etal-2025-domain}. 
Tables~\ref{tab:data} and \ref{tab:model} summarize diverse evaluation approaches such as FLORES for machine translation \citep{goyal-etal-2022-flores, Gala2023}.
% While some models excel in zero-shot generalization \citep{Huzaifah2024,j-etal-2024-romansetu}, others require extensive finetuning for low-resource languages \citep{Nath2023, Chowdhury2022, Singh2024, Dalal2024}.
NER \citep{Venkatesh2022, Khemchandani2021, j-etal-2024-romansetu} and sentiment analysis \citep{Hande2022,Singh2024} usually include accuracy,
F1-score, precision, and recall.
% For example, CPT-R and IFT-R \citep{} with 7B parameters achieve strong zero-shot accuracy (72.5\% on sentiment analysis, 78.2\% F1 on NER) due to extensive pretraining.
MRR (Mean Reciprocal Rank) and NDCG (Normalized Discounted Cumulative Gain) are common evaluation approaches for retrieval and ranking tasks \citep{Haq2023}.
BLEU, ROUGE, METEOR, and human evaluations are standard metrics for generation tasks, such as summarization, machine translation, and question answering \citep{Daisy2023, Rajpoot2024, Gala2023}. 
Recent new metrics such as COMET \citep{rei-etal-2020-comet}, phonetic-aware metrics like PhoBLEU \citep{arora-etal-2023-comix}, SPBLEU \citep{alam-anastasopoulos-2025-large}, and chrF++ \citep{popovic-2017-chrf} complement existing ones \citep{Costajuss2024ScalingNM, gajakos-etal-2024-setu}. Overall, current evaluation relies heavily on English-centric benchmarks and metrics (BLEU, F1, etc. ), which can misrepresent true performance on South Asian languages and thus motivate the need for region-specific evaluation frameworks.

\section{Trends and Challenges}
\label{sec:trends_and_challenges}

Building on the contributions reviewed in the previous sections, we now synthesize emerging patterns and persisting challenges. 

\paragraph{Data Scarcity and Quality Issues}
for low-resource languages affect model generalizability and applicability \citep{Gala2023}.
Existing resources, especially small datasets, are often domain-specific (e.g., government or political) due to limited digital content and copyright restrictions, and may potentially introduce cultural or political biases in downstream applications \citep{Gain2022, Ali2024, Urlana2023, Kumar2024}. 
The lack of gold-annotated resources complicates tasks, such as co-reference resolution \citep{Mishra2024}, and the rapidly evolving online discourse hurts model long-term sustainability \citep{Bandarkar2024, Kumaresan2024}.

\begin{table}
  \centering
  \renewcommand{\arraystretch}{1} % Adjust row spacing
  \setlength{\tabcolsep}{6pt}       % Adjust column spacing
  \small
  \begin{tabular}{>{\arraybackslash}p{1.8cm} 
                    >{\arraybackslash}p{5cm}}
    \hline
    \textbf{Challenge} & 
        \textbf{Example} \\
    \hline
    POS Tagging Inconsistency & ``\raisebox{-0.5ex}{\includegraphics[height=1.8ex]{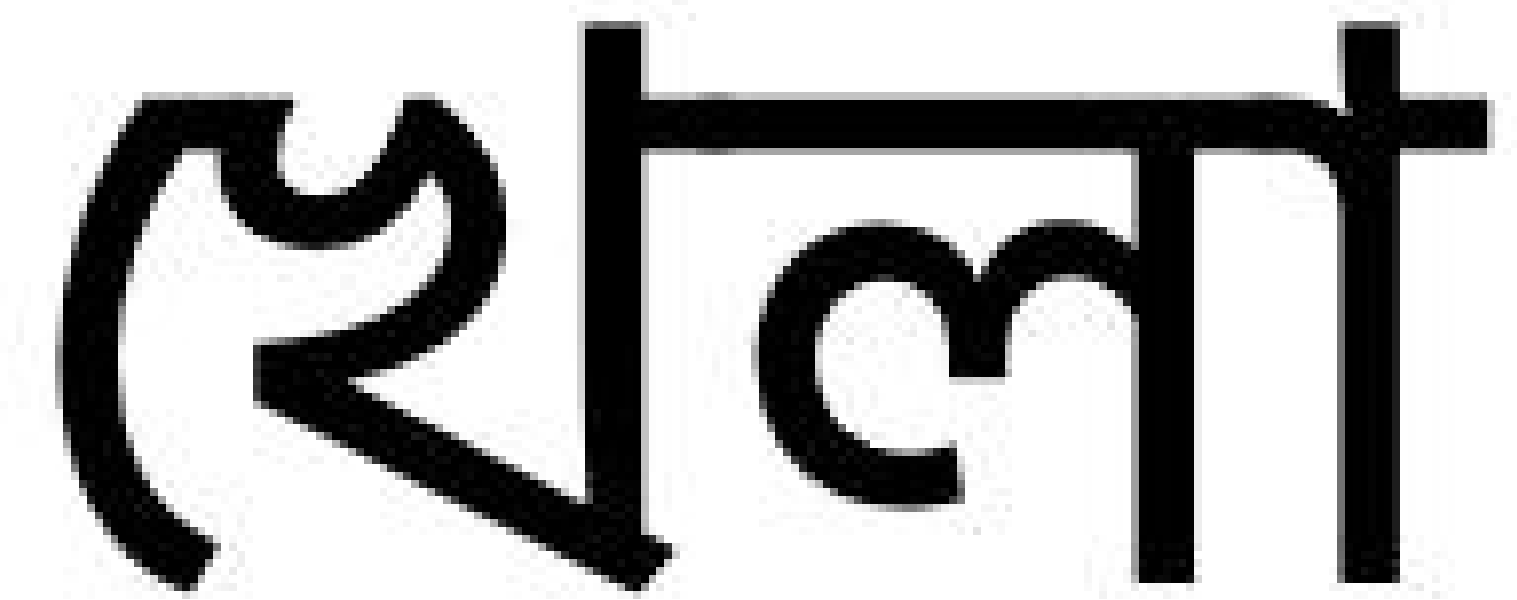}}'' should be tagged as NOUN in ``\raisebox{-0.75ex}{\includegraphics[height=2.3ex]{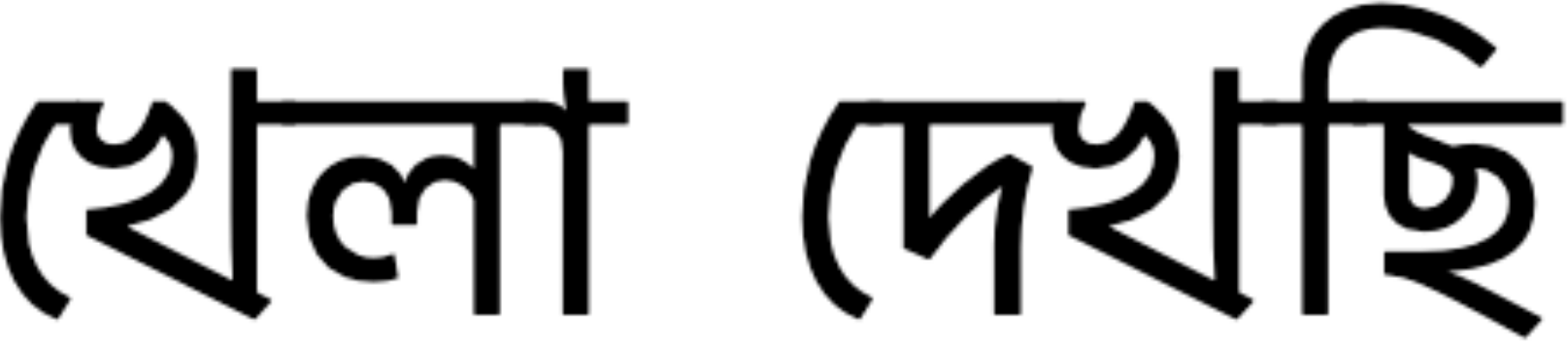}}'' (I am watching a game) and VERB in ``\raisebox{-0.65ex}{\includegraphics[height=2.3ex]{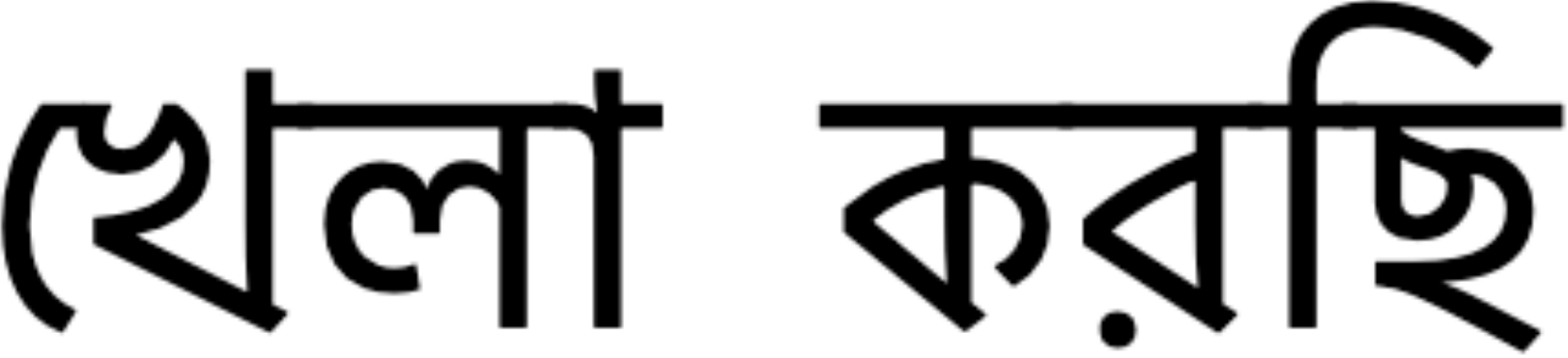}}'' (I am playing)\\
    \hline
    Lexical Variability & Bengali (India): ``\raisebox{0ex}{\includegraphics[height=1.8ex]{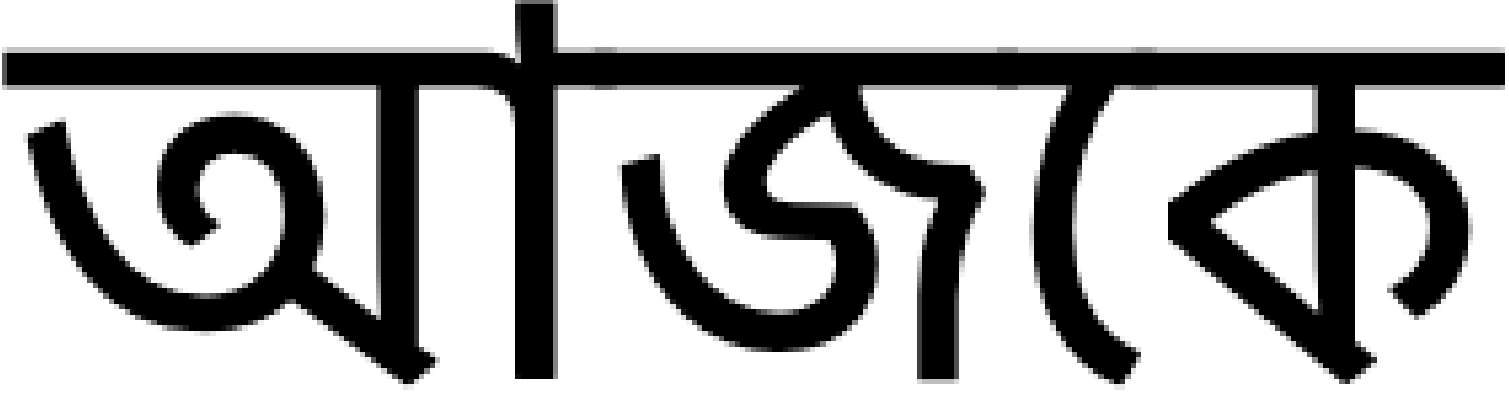}}'' (today); Bengali (Bangladesh): ``\raisebox{0ex}{\includegraphics[height=1.8ex]{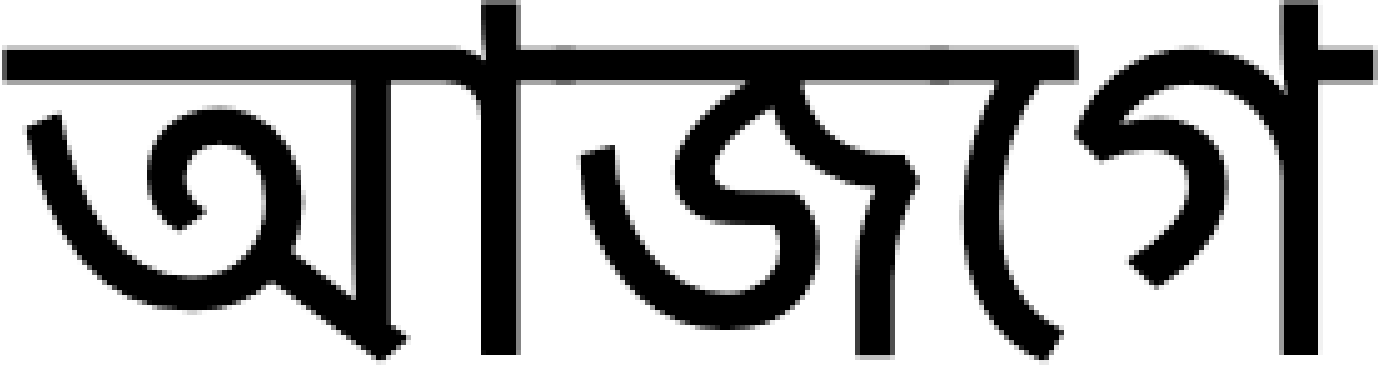}}'' (today)\\
    \hline
    Diglossia & ``Where are you going?'' in Literary Tamil: ``\raisebox{-0.5ex}{\includegraphics[height=2.5ex]{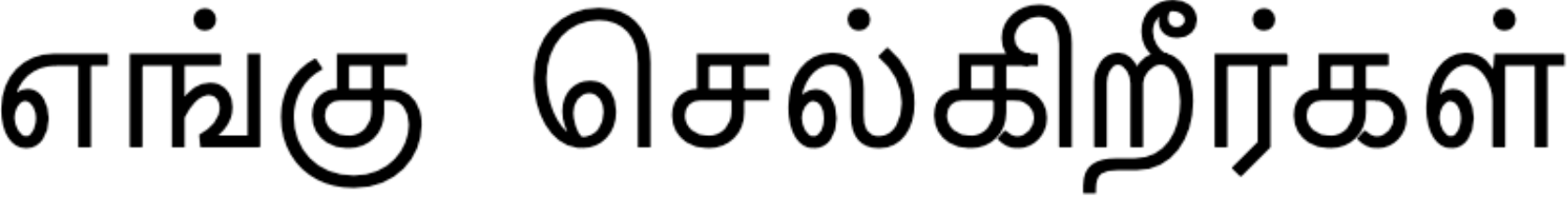}}''; Spoken Tamil: ``\raisebox{-0.5ex}{\includegraphics[height=2.5ex]{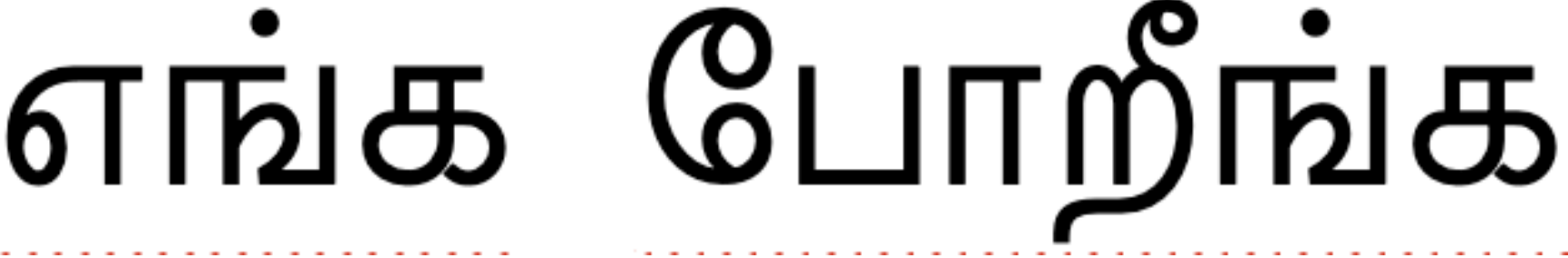}}'' \\
    \hline
    Romanization & Hindi: ``I am fine'' can be romanized as ``main theek hoon'' or ``mai thik hu''\\
    \hline
    Morphological Segmentation & ``\raisebox{-0.5ex}{\includegraphics[height=2.5ex]{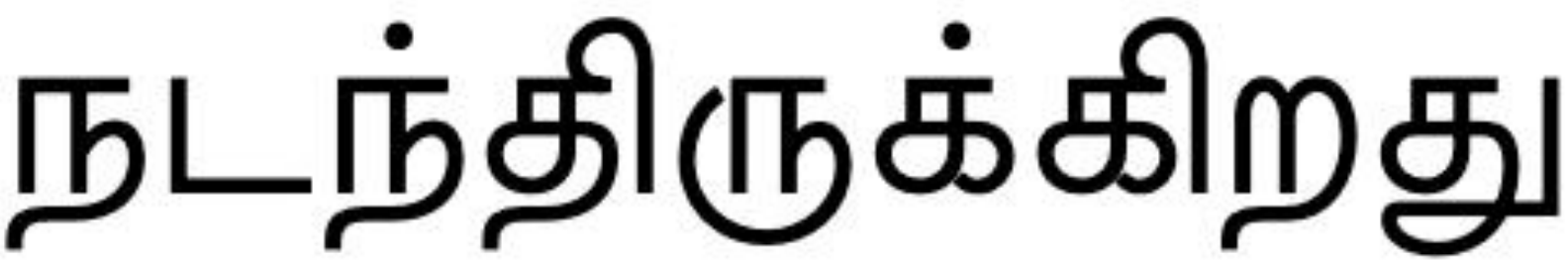}}'' (nadanthirukirathu, ``has happened'') can be broken into [``\raisebox{-0.5ex}{\includegraphics[height=2.5ex]{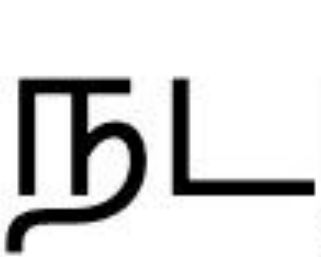}}'' (nada, ``walk'') + ``\raisebox{-0.5ex}{\includegraphics[height=2.5ex]{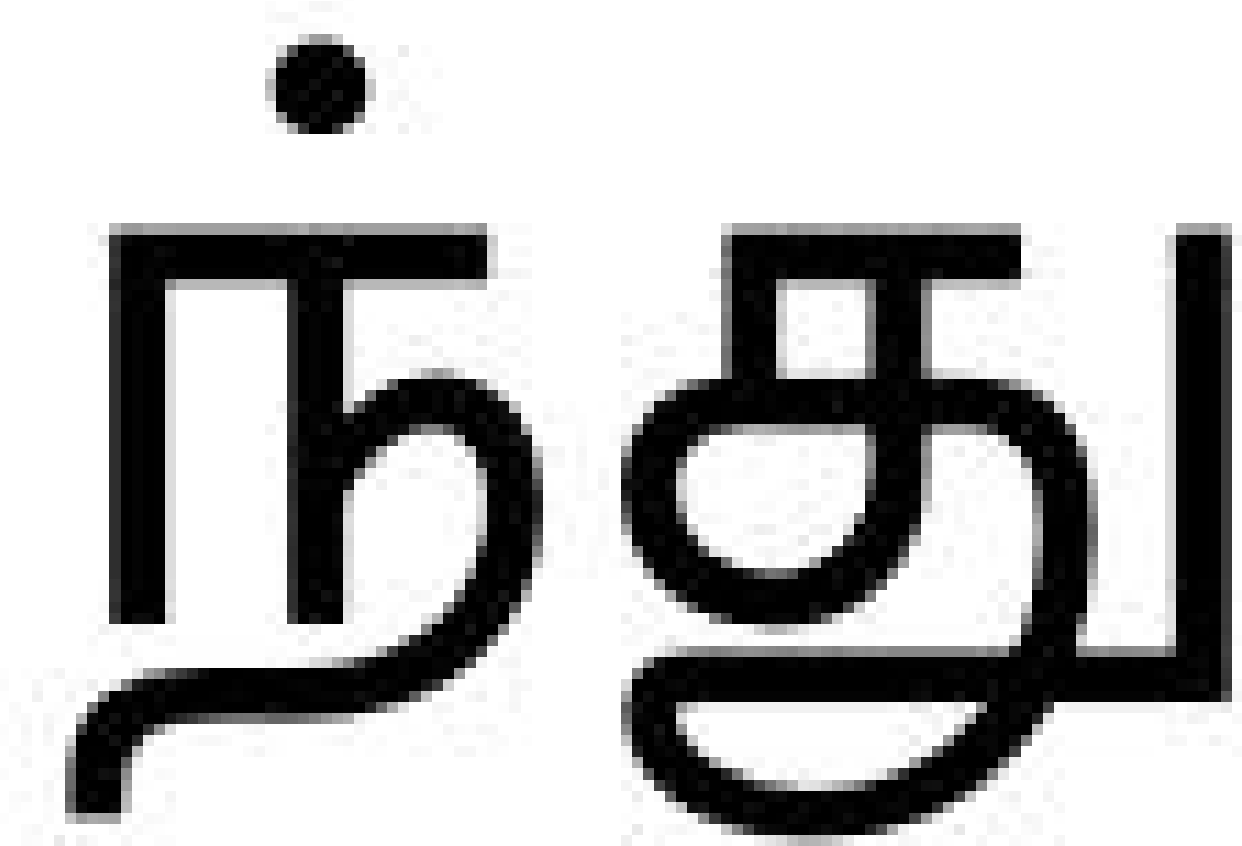}}'' (nthu, past suffix) + ``\raisebox{-0.5ex}{\includegraphics[height=2.5ex]{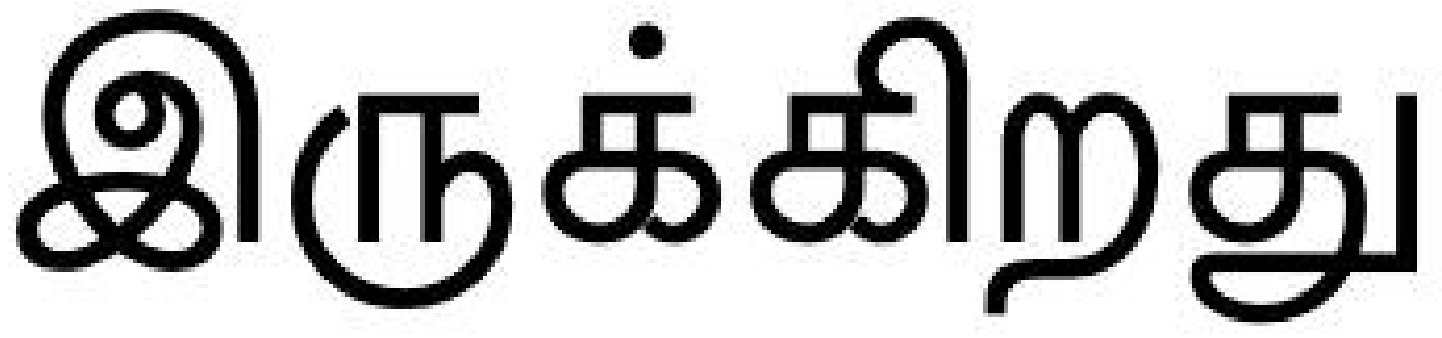}}'' (irukirathu, auxiliary verb) \\
    \hline
    Code mixing & Hinglish: ``Mujhe ek idea aaya'' (I have an idea) \\
    \end{tabular}
  \caption{\label{tab:challenges}
    Linguistic Challenges in Low-Resource South Asian Languages for NLP  
  }
\end{table}

Non-standardized transliteration and representation of South Asian languages introduce biases as annotators often rely on phonetic judgment \citep{Baruah2024}. 
\citet{Bhattacharjee2024} noted inconsistencies in language identification and translation quality due to style and dialect differences within translations and translated text, which are common as for missing human re-verification \citep{Hasan2024}. 
Also, datasets translated from English to a South Asian language can be culturally misaligned \citep{Das2024}.
For culturally nuanced languages \citep{arora2024calmqa}, the requirement for proficient annotators restricts the scalability of data collection efforts.
Biases from human annotators' varying interpretation and background can harm sensitive tasks like hate speech detection \citep{Kumaresan2024}.

Further, certain data exhibit class imbalances, leading to bias toward majority classes; solutions such as cost-sensitive learning and oversampling have been proposed \citep{K2024} but not examined.
Languages exhibiting diglossia need additional efforts as literary text cannot be used for tasks in all settings \citep{Prasanna2024}. 
Limited computing resources further restrict improvements in the curation of high-quality datasets \citep{Philip2021}.

\paragraph{Transliteration and Tokenization Inconsistencies} reduce generalizability of multilingual models on code-mixed languages, such as Hinglish, Tanglish, and Romanized Bengali \citep{Narayanan2024, 10.1145/3695766}. 
Models often learn script-dependent embeddings, which limits cross-script generalization \citep{Koehn2024}.
For example, transliteration ambiguity can easily affect speech-text alignment in ASR models \citep{Ramesh2023}.

Existing tokenization strategies such as Byte-Pair Encoding (BPE)~\citep{gage1994new} and WordPiece~\citep{devlin2019bert} frequently fragment morphologically rich words in Dravidian and Indo-Aryan languages, leading to over-segmentation and loss of meaning \citep{Wang2024}. 
Similarly, agglutinative languages like Tamil and Manipuri form complex word structures that are inconsistently tokenized, affecting syntactic parsing and NMT \citep{Narayanan2024}.
For extremely low-resource languages, pre-trained tokenizers \citep{Kumar2024} fail to adapt effectively as they fragment words into multiple sub-word tokens, sometimes even individual characters, introducing noise to tasks like POS tagging.

Morphological segmentation is particularly challenging for Dravidian languages as words are formed by adding multiple suffixes \citep{Narayanan2024}.
Hindi, Assamese, \& Bengali exhibit different, complex inflectional systems complicating parsing \citep{Chowdhury2018,Nath2023}. 
Most Indo-Aryan languages rely on dependent vowel signs (matras) \& nasalization markers, where BERT tokenizers often split them incorrectly \citep{doddapaneni-etal-2023-towards} and cause ambiguities \citep{Maskey2022}. 
For instance, the word ``\raisebox{-0.75ex}{\includegraphics[height=2.3ex]{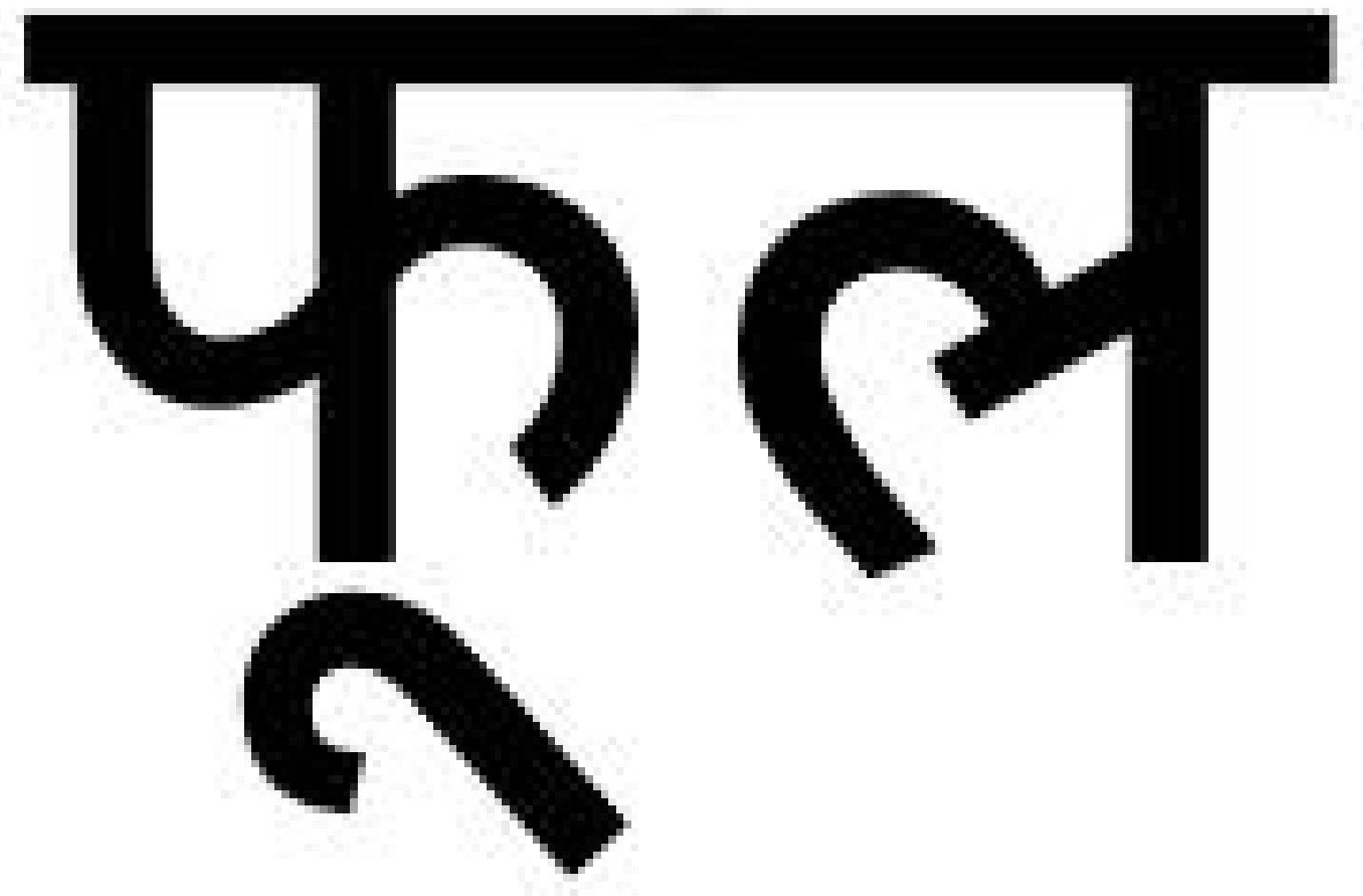}}'' (Flower) can be incorrectly tokenized as ``\includegraphics[height=1.5ex]{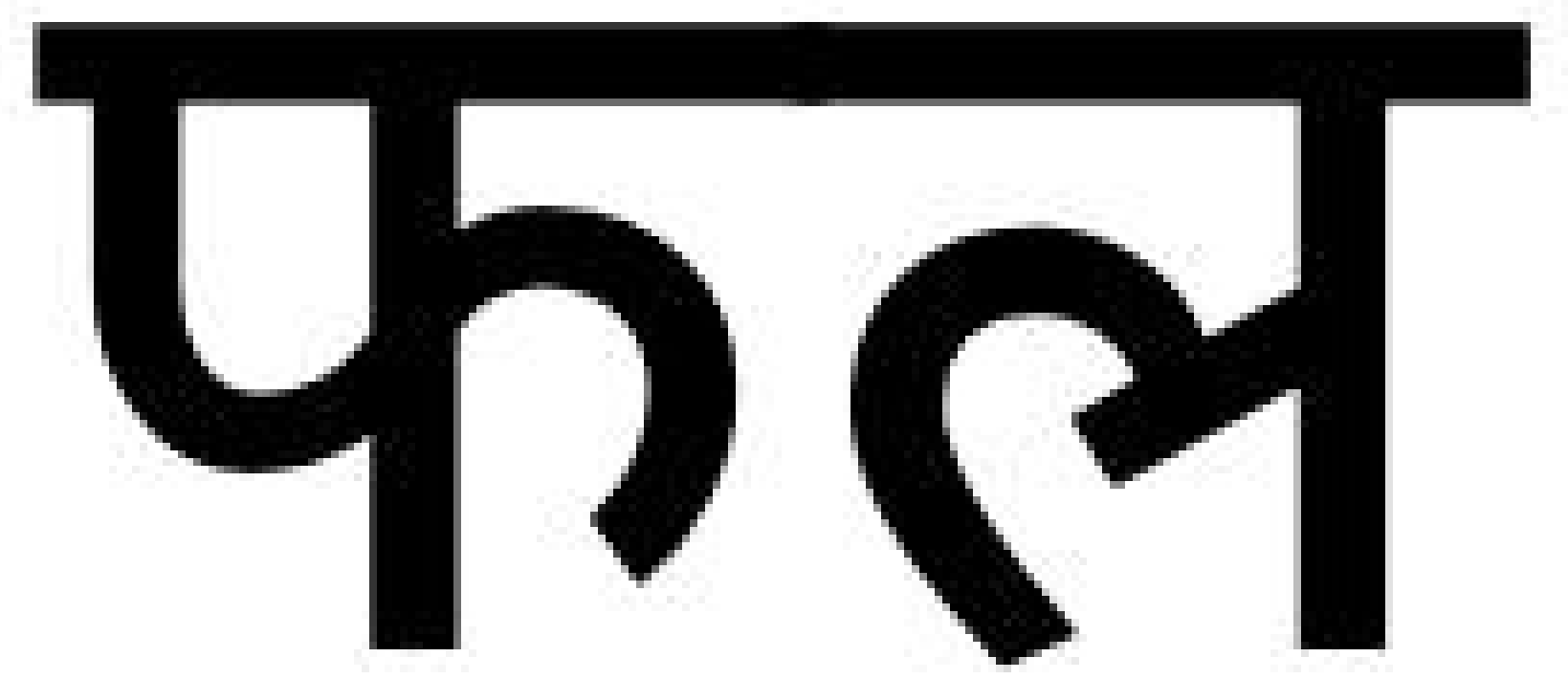}'' (Fruit). 
Assamese possesses unique sound patterns \& alveolar stops, showing the tokenization complexity \citep{Nath2023}.  
Besides structural differences, administrative vocabulary include Persian-origin words like ``farman'' (order), alongside English-origin terms \citep{Pramodya2023}. 
% Additionally, Lambani, spoken by a nomadic tribe, shares linguistic features with multiple Southern Indian languages, an evident of tribal migration \citep{Chowdhury2022}. 

\paragraph{Code mixing, Diglossia, and Ambiguity} are highly domain-dependent issues and can integrate English letters, words, or phrases, such as Hinglish/Tanglish \citep{Das2024}.
Diglossia shows substantial differences in speaking and writing. 
For example, Literary Tamil retains its formal vocabulary, but spoken Tamil incorporates loanwords and phonetic simplifications \citep{Prasanna2024}.
Additionally, polysemy and contextual ambiguities can fail many models on tasks like NER \citep{Bhatt2022}. 
For example, Indic languages do not typically capitalize proper nouns, making it difficult to distinguish named entities from common words \citep{Philip2021}; ``Hindustan'' (\raisebox{-0.75ex}{\includegraphics[height=2.5ex]{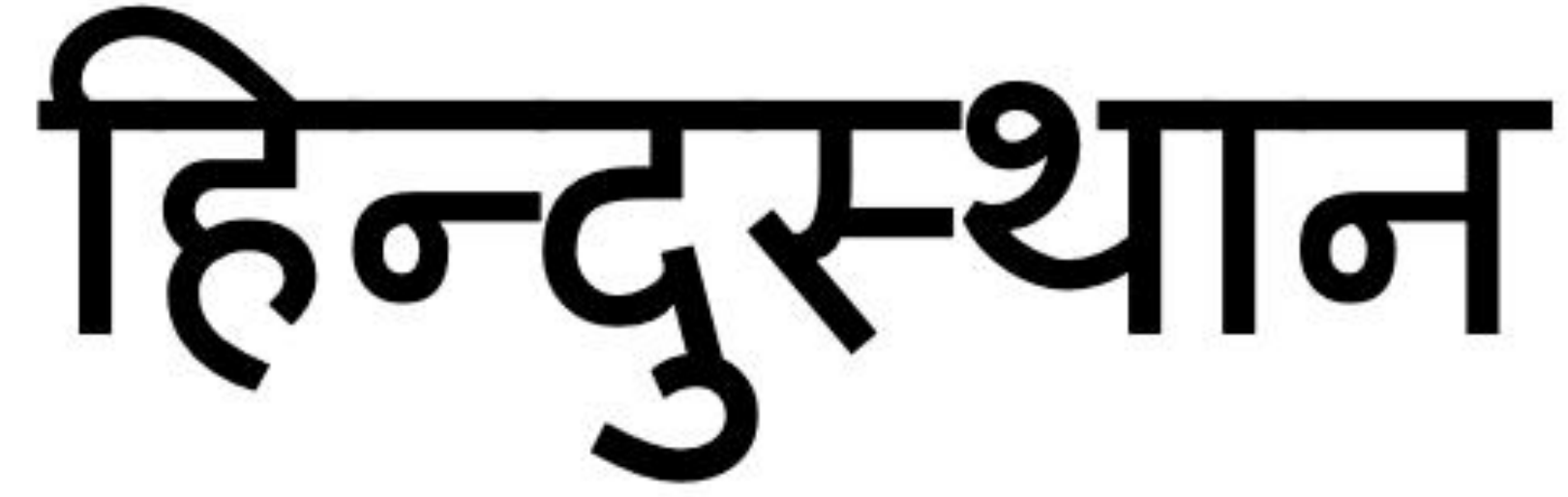}}) can refer to a location, a person, or an organization \citep{Mishra2024}. 
Many languages are grammatically gendered, even inanimate objects being referred to with gendered pronouns \citep{Ramesh2023}.

\paragraph{Dialect Variations and Continua} are common issues in South Asian corpus development as most studies consider a single standard variety. Recent efforts have started addressing this by creating dialect-specific resources \citep{Kumar2024, chowdhury-etal-2025-chatgaiyyaalap, khandaker2024bridging, alam-etal-2024-codet}.
For example, \citet{bafna-etal-2022-combining} curated HinDialect, a folk-song corpus covering 26 Hindi-related dialects; and VACASPATI \citep{bhattacharyya-etal-2023-vacaspati} compiles 115M Bengali literature sentences sampled across West Bengal and Bangladesh to capture regional lexical differences.
Several studies incorporated dialectal cues into models:
AxomiyaBERTa \citep{Nath2023} includes phonological signals via an attention network; \citet{alam-anastasopoulos-2025-large} utilized LoRA \citep{hu2021loralowrankadaptationlarge} to achieve dialectal normalization and translation across South Asian dialects with limited supervision.

However, existing studies show that performance is lower on underrepresented dialects compared to common varieties, which reflects biases in data coverage.
Annotation and orthography for dialectal text are inconsistent––many informal dialects lack standardization and the boundary between ``dialect'' and ``standard'' is often arbitrary \citep{chipsal-ws-2025-1}.
Data frequently conflate dialectal variants with the standard language, while current benchmarks rarely consider these variants.
Most multilingual benchmarks only cover a few dominant languages, so dialectal evaluations are missing.
CHiPSAL and recent shared tasks (e.g., NLU of Devanagari Script Languages) have started to address this by building annotated dialectal corpora \citep{chipsal-ws-2025-1}.
Together, these findings show that dialect-specific corpora and evaluation benchmarks are essential to avoid biasing models toward standard varieties.

\paragraph{LLM Alignment and Reasoning Tasks}
Current LLM benchmarks of South Asian languages suffer with very limited coverage.
For example, the MMLU-ProX covers 13 languages (e.g., Hindi, Bengali) but omits many others such as Tamil, Marathi, \& Kannada \citep{xuan2025mmlu}.
Even broader tests like Global-MMLU span multiple languages ( e.g., Hindi, Telugu, Nepali, etc.) \citep{singh2024global}, yet these datasets were generated by translating English questions.
This leads to cultural mismatch.
Many MMLU \citep{hendrycksmeasuring} questions (e.g., US History, Law) are Western-specific and thus irrelevant in South Asia; \& the translation introduces artifacts that distort evaluation \citep{kadiyala2025improving}.
\citet{ghosh2025multilingualmindsurvey} show that Hindi, the most spoken language in the region, is only represented in 5 multilingual reasoning corpora.

Recent work on cultural and value alignment (CultureLLM) fine-tunes LLMs on global survey data; however, such efforts test broad value judgments rather than deep reasoning in vernacular settings \citep{li2024culturellm}.
For example, \citet{chiu2024culturalbench} covers Bangladesh, India, Nepal, and Pakistan, but the corpus only focuses on trivia/ etiquette and not cultural knowledge in the low-resourced languages spoken in the regions.
In practice, South Asian languages are severely underrepresented in reasoning \& alignment tasks with cultural considerations.

\paragraph{Standard evaluation benchmarks} exist, but gaps have remained in evaluating multilingual models of South Asian language options, distributional balances, and NLP task diversities.
Fine-tuned multilingual models often overfit high-resource regional languages (e.g., Hindi), leading to degraded performance on lower-resource languages \citep{Pal2024}.
Catastrophic forgetting happens when adapting models to new languages or tasks, such as in LoRA and adapter-based finetuning \citep{Nag2024}. 
Phonetic variation across dialects within the same language family (e.g., Bengali \& Assamese) results in inconsistencies in phoneme-based word embeddings \citep{Arif2024}.
Tibeto-Burman \& Austroasiatic evaluation data are almost non-existent and most studies for very low-resourced languages use manually curated datasets \citep{Dalal2024, Chowdhury2022}.

Model evaluation from our collected studies generally rely on English-origin benchmarks in Table~\ref{tab:model}, which can misinterpret model performance \citep{Haq2023}. 
\citet{das-etal-2025-investigating} mentions biases in back-translated datasets cause skewed results, compromising model evaluation across languages. 
For nuanced tasks (e.g., paragraph-level translation), sentence-level evaluation methods may not be sufficient \citep{E2023, Hasan2024}. \citet{mukherjee-etal-2025-evaluating} suggests LLM-based evaluation in the text style transfer task correlates better with human judgment than existing automatic metrics on Hindi and Bengali.
Indeed, without culturally relevant \& task-specific benchmarks, evaluations fail to interpret performance precisely, especially for languages with rich structural/cultural variations \citep{Vashishtha2023}.
% Furthermore, critical dimensions of bias—such as caste and ethnicity—are not widely explored in lower-resourced languages.\\
% FLORES-101 \& FLORES-200 is comprehensive but lacks sufficient representation for code-mixed and dialectal variations \citep{Gala2023}. Multimodal MT efforts highlight a need for image-aware benchmarks \citep{Rajpoot2024}.

\subsection{Multilingual Resources vs South Asian-Specific Efforts}
Broad multilingual resources are attracting more attentions in the NLP communities, such as two recent workshops for South Asian languages \citep{chipsal-ws-2025-1, indonlp-ws-2025-1}.
XNLI benchmark extends English NLI to 14 languages (including Urdu) \citep{conneau-etal-2018-xnli}, and XCOPA provides commonsense reasoning examples in 11 languages \citep{ponti-etal-2020-xcopa}. Similarly, models such as XGLM-7.5B included major South Asian languages \citep{lin2022fewshotlearningmultilinguallanguage}, and new corpora like Glot500 \citep{imanigooghari-etal-2023-glot500} and MaLA-500 \citep{lin2024mala} included over 500 languages. These resources bring valuable South Asian language coverage for cross-lingual evaluation.
However, they rely on general-domain and synthetic data, which can overlook region-specific linguistic and cultural features.
For instance, even XGLM's balanced training includes only approximately 3.4B Hindi tokens versus 803B English, while XCOPA only covers a single Indic language.

Recent efforts explicitly address resource gaps. 
For example, IndicLLMSuite provides 251B tokens of pretraining and 74.8M instruction-response pair data across 22 Indian languages \citep{Khan_2024}, INDIC-MARCO provides MS MARCO-style retrieval queries translated into 11 Indian languages \citep{Haq2023}, 
BPCC parallel corpus contains 230M English-Indic sentence pairs covering 22 Indic languages \citep{Gala2023}, and TransMuCoRes is a coreference resolution data of 31 South Asian languages \citep{Mishra2024}. 
These initiatives incorporate regional linguistic structures (e.g., scripts, complex morphology) and cultural context beyond generic multilingual resources.

Challenges are endless.
Many cross-lingual approaches depend on back translation, introducing new bias and noise and suffering on code-switch (e.g. Hindi-English) issues \citep{raja2025parallel, conneau-etal-2018-xnli}.
%(for instance, prior to very recent efforts, the largest available Nepali corpus was around 11GB in size \citep{thapa-etal-2025-development})
Standard metrics may fail on region-specific phenomena \citep{Mishra2024} among Indic languages. 
These persistent gaps underscore the necessity of region-specific research to ensure equitable and diverse NLP advancements.

\section{Conclusion}

In this study, we provide comprehensive synthesis and analysis of recent NLP advances on low-resourced languages in South Asia.
Our work examines persisting challenges at every stage of resource development––uneven representation in multilingual corpora, model availability, multilingual tuning, and evaluation benchmarks.
While a few languages have received more attention, challenges remain in collecting and processing data and adapting models to specific orthographies.
Moreover, existing evaluation metrics fall short due to a lack of script- and task-specific benchmarks, as well as overlooked sociocultural biases.
We present model tuning guidelines that reflect current limitations of South Asian NLP, calling for South Asian-specific frameworks and script-aware model adaptation.
We include our future envisions in Appendix~\ref{subsec:future}.
We expect this study can encourage broader participation in advancing further research of low-resource languages in South Asia.

\section*{Acknowledgment}
The authors thank anonymous reviewers for their insightful feedback.
This work has been partially supported by the National Science Foundation (NSF) CNS-2318210~\citep{sharif2025ITIGER}.

\section*{Limitations}

Research and development of resources for South Asian languages have been steadily advancing. Significant progress has been made in multilingual datasets and modeling, and many advancements in high-resource languages are now being adapted for low-resource South Asian languages. Since we aimed for a thorough and balanced analysis, below are some key limitations and certain measures we took to address them.

\begin{itemize}
\item Enumerating all studies on low-resource South Asian languages is challenging, as research is dispersed across multiple venues. Many studies are not indexed in the ACL Anthology. During the retrieval stage, we conducted an extensive search across various sources, such as Google Scholar and Semantic Scholar, and have cross-referenced key papers to ensure proper coverage.

\item Identifying relevant studies is complicated due to inconsistent terminology. Papers often use non-standard or domain-specific keywords to describe work on low-resource languages. For instance, some studies refer to `low-resource languages,' while others use `under-resourced languages,' `resource-scarce languages,' or `marginalized languages.' To account for this, we have tested multiple keyword variations and have manually reviewed the related work sections of key papers to identify additional references.

\item Some studies on extremely low-resource languages remain inaccessible because they are published in regional or less widely-indexed journals. We have, to our best efforts, included such publications by searching sources outside of major repositories, especially for Tibeto-Burman and Iranian languages. Future work could benefit from engagement with regional scholars and institutions to access non-digitized resources.
\end{itemize}

% \section*{Acknowledgments}

%\bibliography{custom}
\bibliography{export}

\appendix

\section{Appendix}
\label{sec:appendix}

% This is an appendix.

% what techniques could further help with the existing studies.
\subsection{Study Retrieval and Selection Methodology}
\label{sec:appendix-retrieval}

To identify relevant work on natural language processing for South Asian languages, we conducted an exhaustive literature review led independently by the two authors. 

We ran systematic keyword queries combining South Asian language names (e.g. Hindi, Urdu, Bengali, etc.), region-specific words (e.g., ``Indic'', ``South Asian'', ``Low-Resource Languages''), along with task-specific keywords (e.g., ``Machine Translation'', ``Named Entity Recognition'', ``Sentiment Analysis'', ``Multilingual Pretraining'') across major databases (ACL Anthology, Semantic Scholar, and Google Scholar). This process retrieved over 1,000 initial papers. We then removed duplicates and applied inclusion criteria to focus the review: (a) study of at least one South Asian language with a speaker population $\ge$1 million, (b) use of neural or transformer-based models (e.g., BERT, mBART, T5, GPT), and (c) publication year 2020 or later. After filtering on these criteria, 188 papers remained for full analysis.

All authors independently read and annotated all 188 papers. For each paper, we recorded detailed metadata and qualitative observations using an iteratively-developed structured coding template. Disagreements in coding were resolved through discussion until consensus was reached. The annotation template included both structured metadata (for example: language(s) studied, NLP task, model architecture or family, dataset size, year, and publication venue) and emergent, inductive tags capturing noted phenomena. Examples of inductive tags include transliteration handling, dialectal variation, data scarcity, or evaluation gaps, which were added to the template as they were discovered during reading. These were added as qualitative codes and grouped into higher-order themes.

To ensure coverage of less widely reported research, we searched beyond mainstream venues using citation tracking to identify less accessible research from under-indexed sources. This included work from regional conferences like Technology Journal of Artificial Intelligence and Data Mining, etc. \citep{Kavehzadeh2022}, and workshops focused on low-resource languages. We also scanned citations of benchmark papers like IndicNLG, TransMuCoRes, and BPCC to identify follow-up work not indexed in ACL Anthology. 

We prioritized the inclusion of languages with over 1 million speakers. This allowed us to include both high-resource languages like Hindi and Bengali, as well as low-resource and often overlooked ones such as Manipuri, Balochi, Santali, and Tulu. As discussed in Figure \ref{fig:language_distribution} and Section \ref{sec:language_resources}, the observed imbalance in dataset and model availability reflects publication patterns, not retrieval bias. 

Themes for Sections 3 and 4 were identified inductively by synthesizing recurring patterns across the annotated data. As we reviewed papers, we documented recurrent patterns, gaps, and methodological approaches, which were then grouped into cohesive sections based on relevance to ongoing challenges in South Asian NLP.

% For transparency, we include a summary statistics table (Appendix A, Table 1) breaking down the 188 papers by publication year, venue, and language family. Finally, in the spirit of reproducible research, we open-source the complete list of reviewed papers (with annotations) alongside this survey. This allows others to examine our dataset in detail.

\subsection{Open Challenges and Future Work}
\label{subsec:future}

Building on our survey findings, we outline several forward-looking directions to guide future NLP research for South Asian languages. 

\paragraph{Code-Mixing Beyond Major Language Pairs}
Code-mixing is pervasive in South Asian communication \citep{Huzaifah2024}, yet most available corpora focus on English-Hindi or English-Tamil interactions. We encourage future work to expand toward less-resourced combinations, such as Assamese-Bodo or Hindi-Magahi, and trilingual mixing patterns. Studying the sociolinguistic contexts in which switching occurs (e.g., informal communication, shifts in topic, regional broadcasts) can inform models that generalize better to multilingual discourse. This is particularly relevant for applications like dialogue agents and education technology, where switching is frequent.

\paragraph{Leveraging Bilingualism and Linguistic Proximity for Parallel Data Creation}
Given the high rates of bilingualism in South Asia \citep{bhatia2006bilingualism}, parallel data can be efficiently constructed by pairing low-resource languages with regionally-dominant but better-resourced ones like Hindi, Tamil, or Urdu. We encourage community-driven data collection efforts that take advantage of such speaker fluency. Translation pivots using English–Hindi or English–Tamil models \citep{Khan_2024, Gala2023} can further support indirect transfer. Additionally, our findings on shared scripts and lexical similarity among related languages in Section \ref{sec:language_resources} (e.g., Bhojpuri–Hindi, Assamese–Bengali) suggest promising avenues for cross-lingual data augmentation \citep{Chowdhury2022, Patil2022}.

\paragraph{Bias Mitigation and Inclusive Dataset Design}
As detailed in Section \ref{sec:trends_and_challenges}, our review identifies persistent sociocultural biases in existing resources, ranging from gender and caste under-representation to cultural misalignment in machine-translated data \citep{Bhatt2022, Ramesh2023}, with many datasets relying on translations from English. Very recent work on Nepali-English MT \citep{khadka-bhattarai-2025-gender} also highlights that traditional systems perpetuate gender stereotypes in occupational terms (while GPT-4o demonstrates lower bias and better gender accuracy). However, there are no South-Asian specific large-scale bias evaluation resources. Future work should prioritize participatory dataset development, with native speaker involvement in both content and annotation design. Additionally, targeted efforts are needed to build corpora for languages with scheduled or official status but little NLP presence (e.g., Bodo, Sindhi, Dzongkha, Pashto).

\paragraph{Evaluation Frameworks Tailored to South Asia}
Existing benchmarks rarely capture the linguistic complexity of South Asian languages (e.g., diglossia, agglutination, script multiplicity). Metrics such as BLEU or COMET are often used by default despite them lacking sensitivity to regional variations. We call for the creation of culturally grounded evaluation datasets across tasks like summarization, retrieval, and QA \citep{Philip2021, Kumar2024, pourbahman-etal-2025-elab}, alongside human-in-the-loop assessments in multilingual and code-mixed contexts.

\paragraph{Developing Computationally Efficient NLP Models}

As noted by \citet{Philip2021}, South Asian research institutions often face compute constraints. Future work should prioritize efficient fine-tuning strategies such as adapter-based tuning and LoRA. For example, fine-tuning multilingual LLMs with language-specific instructions \citep{Khan_2024} or leveraging LoRA-based adapters \citep{Huzaifah2024, Singh2024} can yield strong performance with minimal data. Additionally, reasoning and logical inference is being explored in multilingual contexts \citep{ghosh2025multilingualmindsurvey}, but remains under-explored in South Asian NLP. Further research would improve the decision-making capabilities of models catering to South Asian languages.

\paragraph{Script-Robust and Transliteration-Aware Modeling}
South Asian languages often use multiple scripts or informal romanizations. The survey notes that transliterating text into a common script can improve cross-lingual transfer, but current models still suffer from script-specific tokenization issues \citep{Koehn2024}. Recent work such as Nayana \citep{kolavi-etal-2025-nayana} demonstrates that combining synthetic layout-aware data generation with LoRA can enable scalable OCR for 10 Indic languages wihtout requiring annotated corpora.

Future research should focus on script-agnostic modeling: for example, designing multilingual tokenizers or shared subword vocabularies that link Devanagari, Perso-Arabic, and Roman scripts. Modules that automatically transliterate or phonetically encode text (so that Hindi and Urdu versions of the same word align) could boost transfer. Such techniques (training on mixed-script data or using script-independent representations) will help models generalize across writing systems common in South Asia.

\paragraph{Coordinated South Asian Benchmarks and Shared Tasks}
We observe fragmented evaluation across studies, with little standardization. Inspired by initiatives like IndicGLUE \citep{kakwani-etal-2020-indicnlpsuite} and BigScience \citep{akiki2022bigscience}, we propose community-organized shared tasks focused on regionally relevant domains (e.g., healthcare, law, government communication) and languages. These should include multilingual, multi-script benchmarks, standardized metrics, and code-mixed test sets to advance reproducibility and collaboration.

\end{document}